\documentclass[conference]{IEEEtran}
\IEEEoverridecommandlockouts

\usepackage{cite}
\usepackage{amsmath,amssymb,amsfonts}
\usepackage{algorithmic}
\usepackage{graphicx}
\usepackage{textcomp}
\usepackage{xcolor}
\usepackage{multirow}
\usepackage{booktabs}
\def\BibTeX{{\rm B\kern-.05em{\sc i\kern-.025em b}\kern-.08em
    T\kern-.1667em\lower.7ex\hbox{E}\kern-.125emX}}
\begin{document}

\title{MX-Font++: Mixture of Heterogeneous Aggregation Experts for Few-shot Font Generation}

\author{
\IEEEauthorblockN{Weihang Wang*}
\IEEEauthorblockA{\textit{BILIBILI Inc.} \\
Shanghai, China \\
kiren.wwh@outlook.com}
\and
\IEEEauthorblockN{Duolin Sun*}
\IEEEauthorblockA{\textit{BILIBILI Inc.} \\
Shanghai, China \\
sdl2021@mail.ustc.edu.cn}
\and
\IEEEauthorblockN{Jielei Zhang\dag}
\IEEEauthorblockA{\textit{BILIBILI Inc.} \\
Shanghai, China \\
yctmzjl@gmail.com}
\and
\IEEEauthorblockN{Longwen Gao}
\IEEEauthorblockA{\textit{BILIBILI Inc.} \\
Shanghai, China \\
gaolongwen@gmail.com}
\thanks{* Equal Contribution.

\dag \ Corresponding Author.}
}

\maketitle

\begin{abstract}
Few-shot Font Generation (FFG) aims to create new font libraries using limited reference glyphs, with crucial applications in digital accessibility and equity for low-resource languages, especially in multilingual artificial intelligence systems. Although existing methods have shown promising performance, transitioning to unseen characters in low-resource languages remains a significant challenge, especially when font glyphs vary considerably across training sets.
MX-Font considers the content of a character from the perspective of a local component, employing a Mixture of Experts (MoE) approach to adaptively extract the component for better transition. 
However, the lack of a robust feature extractor prevents them from adequately decoupling content and style, leading to sub-optimal generation results. 
To alleviate these problems, we propose Heterogeneous Aggregation Experts (HAE), a powerful feature extraction expert that helps decouple content and style downstream from being able to aggregate information in channel and spatial dimensions. Additionally, we propose a novel content-style homogeneity loss to enhance the untangling. 
Extensive experiments on several datasets demonstrate that our MX-Font++ yields superior visual results in FFG and effectively outperforms state-of-the-art methods. Code and data are available at https://github.com/stephensun11/MXFontpp.

\end{abstract}

\begin{IEEEkeywords}
Few Font Generation, Mixture of Experts, Cross-lingual Font Generation.
\end{IEEEkeywords}

\section{Introduction}
Font generation\cite{tian2016rewrite, tian2017zi2zi, lyu2017auto, azadi2018multi} serves as a crucial technology for digital typography, enabling the creation of visually consistent and culturally appropriate fonts across different writing systems. This is particularly important for addressing the digital divide, as many languages lack adequate font libraries for digital content creation and display. However, traditional font design requires extensive manual effort to create each character, making it impractical for languages with large character sets or limited digital resources. To address this challenge, Few-shot Font Generation (FFG)\cite{xie2024weakly, tang2022few, wang2023cf} aims to generate new font styles with only a small number of sample images\cite{zheng2020jnd, chang2018pairedcyclegan, li2018beautygan, choi2018stargan, liu2019few, tumanyan2022splicing}, dramatically reducing the required manual effort while maintaining typographic consistency.
It is a challenging endeavor in FFG studies\cite{sun2017learning, zhang2018separating, gao2019artistic} to capture the precise patterns and structures of languages with complex strokes and numerous characters, such as Chinese or Arabic.

\begin{figure}[htbp]
    \centering
    \small
    \includegraphics[width=0.8\linewidth]{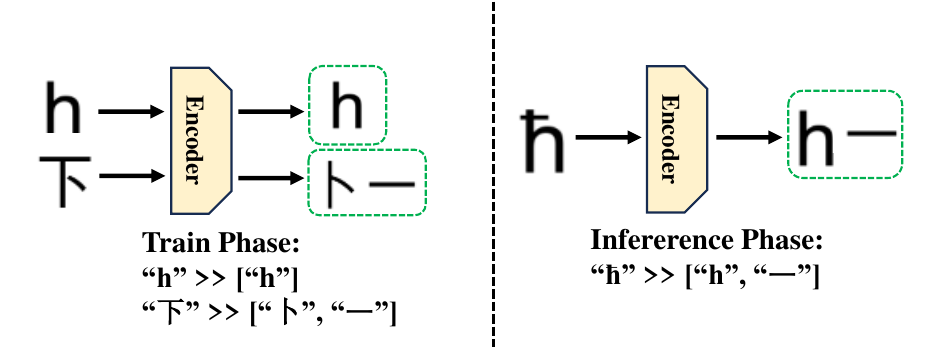}
    \vspace{-10pt} 
    \caption{The encoder adopts an adaptive component allocation process during both training and inference. During the training phase, it enhances the model's ability to predict character components. During inference, when dealing with unseen characters, it adaptively searches for the corresponding set of components in the component library.}
    \label{fig:picture001}
    \vspace{-20pt}
\end{figure}

FFG models \cite{liu2023fonttransformer,bai2024intelligent,park2021multiple} typically comprise a content encoder and a style encoder, which are responsible for extracting content and style features from specified character images, respectively. These features are then fused and decoded to generate character images that embody the desired content and style (i.e., fonts). 
However, the primary challenge lies in decoupling content and style features effectively, as they contain rich and intertwined information. Relying solely on convolutional neural network encoders for their separation often results in sub-optimal feature disentanglement, particularly for complex scripts and cross-lingual scenarios.

Current FFG methods are usually categorized into two types: global information-based methods\cite{sun2017learning, liu2019few, gao2019artistic} and component-based methods\cite{cha2020few, park2021multiple}. Global methods use a content encoder combined with supervised signals (e.g., character classification), while component-based methods decompose a character image into multiple parts and train each part using content and style signals. Due to the glyph complexity of some characters, global methods have difficulty capturing local information and therefore produce sub-optimal results. However, cross-language font generation is particularly challenging and requires models with strong migration capabilities as well as localized attention, which global approaches lack. Experimental evidence suggests that component-based approaches (e.g., MX-Font\cite{park2021multiple}) excel in this regard, even if Latin or Arabic alphabets do not have specific components (e.g., radicals or strokes), presumably because they are more effective at capturing the implicit glyph information inherent in multilingual fonts. 
Specifically, MX-Font\cite{park2021multiple} proposes that a Mixture of Experts (MoE)\cite{jacobs1991adaptive} can be used to capture implicit component information in characters, which is important for texts that lack specific component annotations, as shown in Figure \ref{fig:picture001}. However, their experts rely on a simple convolutional structure that ignores both spatial and channel information in the images. This limitation reduces the model's ability to effectively capture implicit component information, leading to sub-optimal results, especially for resource-poor languages. The impact of this limitation extends to downstream applications such as Scene Text Recognition (STR)\cite{du2022svtr,liao2019scene,wan2020textscanner,xie2024deeptts,xie2024dntextspotter,wan2020vocabulary}, where the quality of generated fonts directly affects recognition performance.

In this paper, we propose MX-Font++, which improves the weak feature extraction capability of MX-Font. Specifically, firstly, we propose Heterogeneous Aggregation Experts (HAE), which possesses the Transformer Encoder structure, and the Attention part adopts the Heterogeneous Aggregation Attention (HAA) module that can aggregate the channel and spatial information so that the model can better separate the content and style features. In addition, in order to ensure the decoupling of the content and style, we propose a new loss function called the content-style homogeneity loss, which increases the heterogeneity between content and style features in latent space. Extensive experiments show that our proposed MX-Font++ significantly outperforms the state-of-the-art methods in FFG. Overall, our work contributes in three main aspects:

1. We propose a method based on Mixture of Heterogeneous Aggregation Experts for FFG, which can better enhance the expert's feature extraction ability.

2. We introduce a novel content-style decoupling loss, namely the content-style homogeneity loss, to address the content-style entanglement problem.

3. Through a series of detailed experiments, the results demonstrate that MX-Font++ achieves state-of-the-art results in the Chinese generation and cross-lingual generation, even for languages with limited resources.

\begin{figure*}[htb]
    \centering
    \small 
    \includegraphics[width=0.9\textwidth]{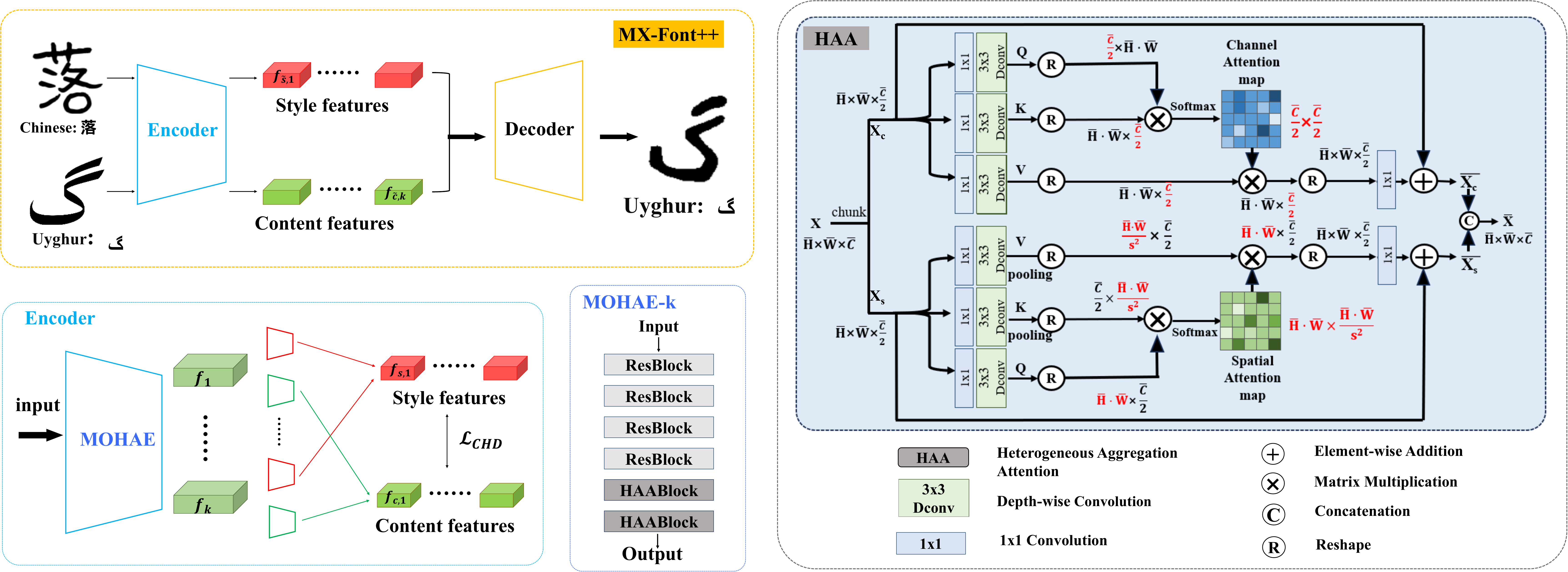}
    \caption{The proposed architecture of MX-Font++ (top left). The overall framework consists of two main parts: encoder and decoder. The encoder part uses our proposed Mixture of Heterogeneous Aggregation Experts (MOHAE) to encode the characters to obtain style and content features (bottom left and bottom half). After that, the style features and content features from different characters are combined and decoded to obtain the final character. MOHAE uses $k$ Heterogeneous Aggregation Experts (HAE) as the base encoder, which is a heterogeneous aggregation encoder architecture that facilitates the decoupling of content and style (right).}
    \label{fig:arch}
\end{figure*}

\section{Method}
To address the issue of generalization to unseen components (e.g., new language systems) in previous component-based methods, MX-Font proposes using a Mixture of Experts (MoE). In this approach, different components are represented by multiple experts, enabling generalization to new language systems. Specifically, MX-Font encodes the image $x$ using $k$ experts \{$E_1,E_2,...,E_k$\} via simple convolutional layers, producing encoded features $f_{i}=E_{i}(x)$, where $E_i$ is the $i$-th expert and $f_i$ is the encoded feature. Two linear layers $W_{i,c}$ and $W_{i,s}$ are then used to obtain content features and style features $f_{c,i}=W_{i,c}^\top f_i, f_{s,i}=W_{i,s}^\top f_i$. Finally, $Cls_c$ and $Cls_f$ supervise the features to ensure the model extracts the corresponding style and content features based on the local components.

The MX-Font\cite{park2021multiple} method uses simple experts and convolutional layers to separate content and style, which affects the final generated image. We propose Heterogeneous Aggregation Experts (HAE) to enhance feature extraction from both channel and spatial perspectives for better separation of content and style. Additionally, we introduce content-style homogeneity loss to help HAE decouple style and content. Finally, we use generator G to combine the content features of the content image with the style features of the style image to generate the final image:
\begin{equation}
    \widetilde{x}=G((f_{s,1}\circ f_{c,1}),\ldots,(f_{s,k}\circ f_{c,k})),
\end{equation}
$\circ$ denotes a concatenate operation. We utilize discriminator $D$ to supervise the generated images. Figure~\ref{fig:arch} describes the specific framework of our model.

\subsection{Heterogeneous Aggregation Experts}
To optimally separate content and style, the feature extraction module must consider both spatial and channel dimensions of the image. Inspired by HAFormer\cite{sun2024haformer}, we use Heterogeneous Aggregation Experts (HAE) to aggregate this information effectively. HAE based on the Transformer Encoder structure, enhances feature extraction. Initially, a simple CNN extracts features from image \(x\): \(z = CNN(x)\). The feature \(z\) undergoes Layer Normalization, passes through the Heterogeneous Aggregation Attention (HAA) layer, and is residually linked to the initial \(z\). Finally, a feed-forward network further transforms the features. 
This process is formulated as:
\begin{align}
    &z' =  z + \text{HAA}(LN(z)), \\
    &f =  z' + \text{FFN}(LN(z')).
\end{align} 
Thus, the combination of these elements permits a more efficient and comprehensive feature extraction.

Our HAA adopts a dual-branch architecture, where one branch handles channel information and the other branch handles spatial information. Finally, we will fuse the outputs of the two branches:
\begin{align}
    z_{s}, z_{c} & = \text{Chunk}(z), \\
    \overline{z} & = \text{Concat}(g_s(z_s), g_c(z_c)).
\end{align}
$z$ is the feature processed by the LN layer, $Chunk$ is separated along the channel, $z_s$ and $z_c$ are spatial and channel features, $g_s$ and $g_c$ are spatial modeling and channel modeling functions, respectively. The architecture of HAA is shown in Figure \ref{fig:arch}. 

In our channel modeling, we use self-attention mechanisms to create attention maps \(A_{c} \in \mathbb{R}^{\frac{\overline{C}}{2} \times \frac{\overline{C}}{2}}\), indicating channel significance. For spatial modeling, we pool \(K\) and \(V\) but not \(Q\) to maintain high-resolution details. The spatial attention map \(A_{s} \in \mathbb{R}^{\overline{H} \cdot \overline{W} \times \frac{\overline{H} \cdot \overline{W}}{s^2}}\) represents regional importance, with \(s\) as the pooling hyper-parameter to reduce computational complexity. Our HAA aggregates channel and spatial information, enabling our MoE encoder to extract more details from character images.

\subsection{Separate Content and Style}
To separate content and style features, we use two classifiers during training to supervise these features, excluding them during inference. The style classifier supervises \(W_s\) by classifying \(f_s\), where the category label of \(f_s\) is the one-hot encoding of its associated font:
\begin{equation}
    \hat{y}_s = \xi_s(f_s)
\end{equation}
\begin{equation}
    f_s = Concat(f_{1,s}, f_{2,s}, ... f_{k,s})
\end{equation}
Existing font generation methods often predict the entire character as the content classifier, which can lead to a large target set and coarse supervision signals. Some methods use stroke information for fine-grained supervision, but this can be labor-intensive for annotating new characters, counteracting the goal of reducing human effort.

We employ component information as supervision signals to achieve fine-grained supervision. The content classifier predicts the component to which each \(f_{i,c}\) belongs, connecting all content features to components. The formulation is:
\begin{equation}
Comp_i = \xi_c(f_{i,c}) 
\end{equation}
Where \(Comp_i\) represents the component contained in \(f_i\), and $\xi_c$ denotes the content classifier. By predicting all components \(\hat{Comp} = \{Comp_1, Comp_2, ..., Comp_k\}\), further comparison with \(Comp_{gt}\) serves as the supervision signal to train \(W_c\).

To further separate content and style, we also propose content-style homogeneity loss to optimize both features from the perspective of Euclidean distance. The formulation is:
\begin{equation}
    \mathcal{L}_{csh} = (\frac{f_s \cdot f_c}{\|f_s\| \|f_c\|} + 1) / 2
\end{equation}
where $f_c$ and $f_s$ represent the content features and the style features output from the encoder.

\section{Experiments}

In this section, we evaluate our MX-Font++ for the Chinese and multilingual font generation. First, we introduce the dataset of our experiments. Then we report and analyze the experimental results of our method and various baselines. We also conduct ablation studies to validate the effects of different components in our method. Extensive experiments demonstrate the superiority of our MX-Font++.

\renewcommand{\arraystretch}{1.5} 
\begin{table}[t]
    \centering
    \caption{Chinese generation performance of our MX-Font++ and other methods on two different settings. 
    }
    \vspace{-8pt}
    \label{tab:chinese_generation}
		\centering 
  \resizebox{\linewidth}{!}{
  \begin{tabular}{c|c|cccccc}
			\toprule
			\multirow{1}{*}{Dataset}&Method
			&SSIM$\uparrow$ &LPIPS$\downarrow$ &FID$\downarrow$ &L1$\downarrow$ &RMSE$\downarrow$ &User$\uparrow$\cr
			\midrule
   
			\multirow{6}{*}{UFSC}
			& MX-Font &0.532&0.289&114.81&8.24&5.46&5.43\cr
			~ & FS-Font  &0.562&0.311&121.07&7.93&5.27&5.03\cr
			~ & CG-GAN &0.594&0.368&117.74&6.71&5.74&4.07\cr
			~ & FontDiffuser &0.617&0.234&113.58&7.25&4.28&8.18\cr
			~ & \textbf{MX-Font++} &\textbf{0.689}&\textbf{0.201}&\textbf{103.94}&\textbf{6.03}&\textbf{3.23}&\textbf{9.09}\cr
			\hline
                
                \multirow{6}*{UFUC}
			& MX-Font &0.468&0.335&118.59&8.34&6.27&6.75\cr
			~ & FS-Font &0.489&0.347&113.27&8.23&5.97&5.90\cr
			~ & CG-GAN &0.454&0.399&112.88&8.79&4.75&1.50\cr
			~ & FontDiffuser &0.456&0.293&112.28&7.96&4.96&8.00\cr
			~ & \textbf{MX-Font++} &\textbf{0.653}&\textbf{0.279}&\textbf{108.37}&\textbf{6.58}&\textbf{3.49}&\textbf{8.80}\cr
			\hline
		\end{tabular}
  }
    \vspace{-0.15in}
\end{table}

\subsection{Dataset and Evaluation Metrics}

Our evaluation includes Chinese and multilingual font generation. \textit{For the Chinese generation}, we use a dataset of 400 fonts and 3346 characters, with 385 fonts and 3200 characters for training. The remaining 15 unseen fonts serve as a test set for all characters, categorized as seen and unseen. \textit{For multilingual generation}, we apply the Chinese model cross-lingually, using training fonts as style input and Unicode fonts as content. We test on Cyrillic and Latin alphabets used by ethnic minorities. 
After generating the fonts, we utilize Synthtiger\cite{yim2021synthtiger} to create a recognition training set of 600,000 samples for each language. We then train a recognition model using SVTR\cite{du2022svtr} architecture and evaluate its performance on our low-resource language STR test dataset.

We evaluate Chinese font generation using L1 loss, RMSE, SSIM, LPIPS\cite{zhang2018unreasonable}, and FID\cite{heusel2017gans}. For multilingual fonts, we use line and character accuracy scores on a test set evaluated by recognition models trained on synthetic data.

Additionally, we conduct a user study to evaluate our generated results. We curate inference results from four state-of-the-art methods and MX-Font++ on Unseen Font Seen Character (UFSC) and Unseen Font Unseen Character (UFUC) datasets, selecting 15 characters from each dataset for testing. 

\begin{table}[t]
    \centering
    \caption{STR performance obtained by training on different models\cite{du2022svtr} using images generated by the current method as training data ("Seq." for sequence accuracy, "Char." for character accuracy). The third row shows the user study.}
    \vspace{-0.1in}
    \resizebox{\linewidth}{!}{
    \label{tab:russia_ocr}
    \begin{tabular}{cc|ccccc}
    \hline
    \multicolumn{2}{c|}{Method} & FS-Font & MX-Font & CG-GAN & FontDiffuser & \textbf{MX-Font++} \\ \hline
    \multirow{2}{*}{SVTR-Small} 
    & Seq. & 0.1536 & 0.2830 & 0.1248 & 0.2627 &\textbf{0.4033} \\ 
    & Char. & 0.4878 & 0.6355 & 0.2858  & 0.5958  & \textbf{0.7519} \\ \hline
    \multirow{2}{*}{SVTR-Large} 
    & Seq. & 0.3052 & 0.4568 & 0.2095 & 0.3773 &\textbf{0.6521} \\ 
    & Char. & 0.6186 & 0.7792 & 0.3858  & 0.6927  & \textbf{0.8564} \\ \hline
    \multicolumn{2}{c|}{User Study} & 4.85 & 5.88 & 1.59 & 7.88 & \textbf{9.37} \\ \hline
    \end{tabular}
    }
    \vspace{-5pt}
\end{table}

\subsection{Comparison with State-Of-The-Art Methods}
We compare our method with four state-of-the-art methods: three GAN-based methods (MX-Font\cite{park2021multiple}, FS-Font\cite{tang2022few}, and CG-GAN\cite{kong2022look}) and one diffusion-based method (FontDiffuser\cite{yang2024fontdiffuser}).

\subsubsection{Quantitative comparison} 
For Chinese font generation, the results in Table \ref{tab:chinese_generation} show that MX-Font++ excels in all evaluation metrics for both seen and unseen characters, outperforming the previous state-of-the-art. For multilingual font generation, we randomly selected 20 characters from low-resource languages and used various methods for cross-lingual generation. Further, we compare our method's Cyrillic recognition accuracy against SOTA methods. Using synthesized samples and SVTR models, MS-Font++ significantly outperformed others. Extended experiments further demonstrated MS-Font++'s superiority and readiness for deployment. Feedback from 20 volunteers indicated that MX-Font++ led significantly in the user study as shown in Table \ref{tab:russia_ocr}.

\begin{figure}[h]
    \vspace{-5pt}
    \centering
    \small
    \includegraphics[width=0.9\linewidth]{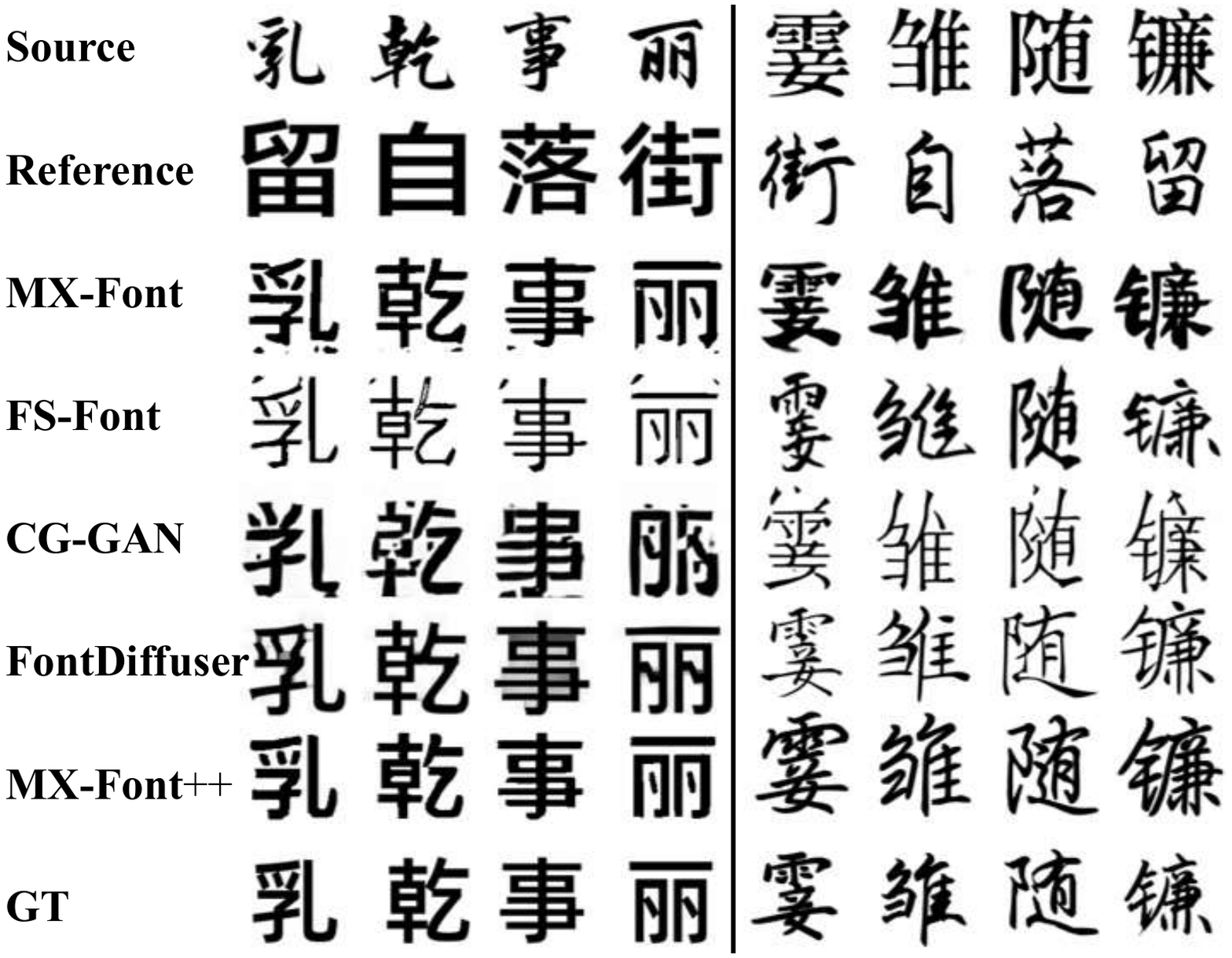}
    \vspace{-6pt} 
    \caption{Samples of Chinese FFG visualization results from different models. The left columns are the UFSC results. The right columns are the UFUC results.}
    \label{fig:vis_zh}
\end{figure}

\subsubsection{Qualitative comparison}
\begin{figure}[t]
    \centering
    \small
    \includegraphics[width=0.9\linewidth]{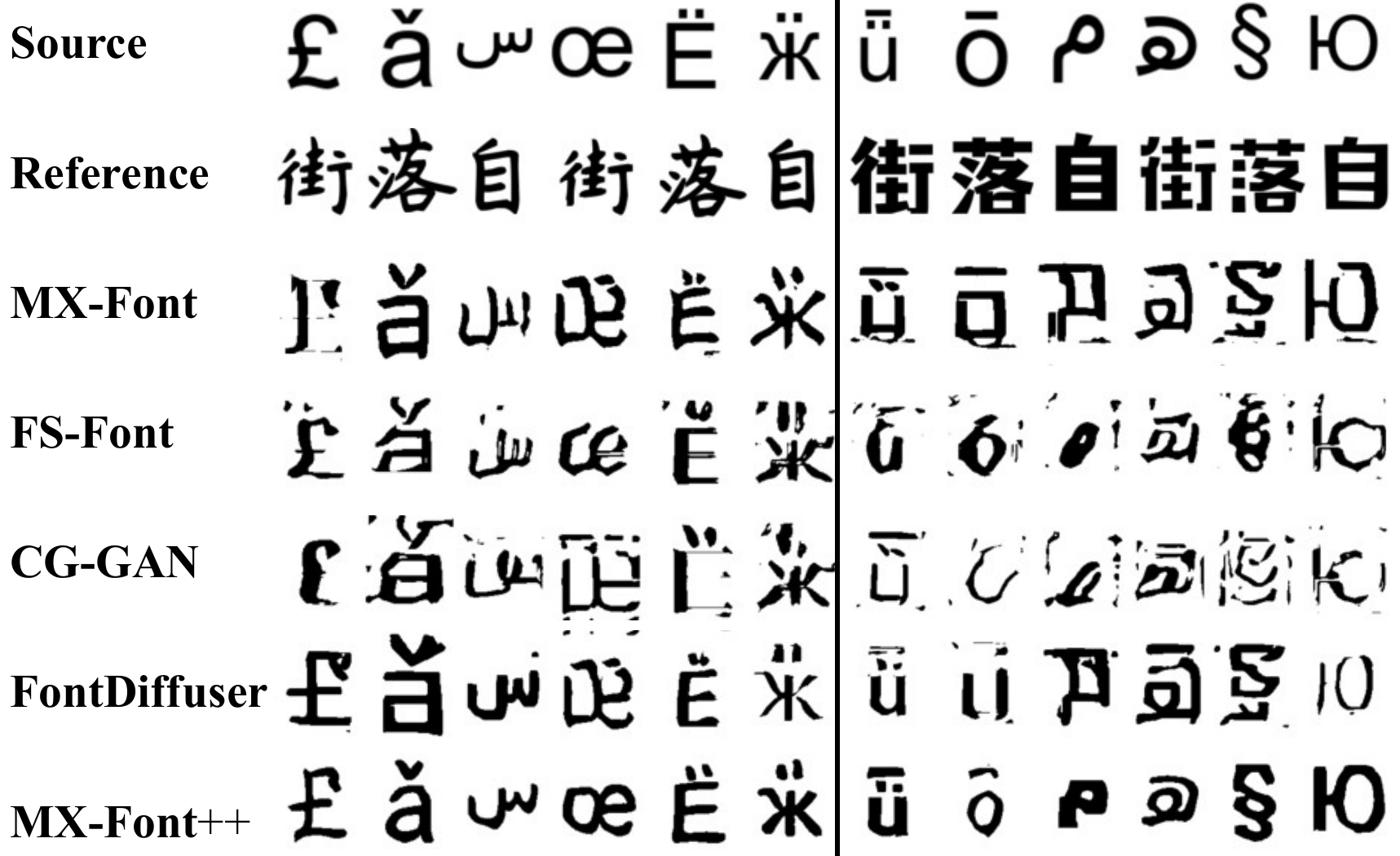}
    \vspace{-6pt} 
    \caption{Samples present cross-lingual FFG results on limited-resource language visualization, obtained from different models. The left columns illustrate the UFSC results, while the right columns depict the UFUC results.}
    \label{fig:vis_multi}
    \vspace{-0.1in}
\end{figure}

In Figure \ref{fig:vis_zh} we present visual representations of both seen and unseen Chinese characters generated by various methods, graphically illustrating the notable disparities identified in the MX-Font++ user study. In addition, we visually compare the low-resource language generation capabilities of different methods by selecting characters from Uyghur and Kazakh languages and generating characters based on reference fonts as shown in Figure \ref{fig:vis_multi}. Our MX-Font++ yields image outputs that surpass those of existing state-of-the-art methods.

\subsection{Ablation Studies}
To validate the effectiveness of the components in our proposed method, we conduct detailed ablation experiments on the HAE and the content-style homogeneity loss. The results are presented in Table~\ref{tab:ablation}.

\begin{table}[htbp]
    \vspace{-0.1in}
	\centering
    \caption{Ablation results of MX-Font++.}
    \label{tab:ablation}
        \vspace{-0.1in}
        \resizebox{\linewidth}{!}{
	\begin{tabular}{c|ccccc|ccccc} \hline
		& \multicolumn{5}{c|}{Unseen Font Seen Character (UFSC)} & \multicolumn{5}{c}{Unseen Font Unseen Character (UFUC)}\\ 
            \hline 
		& SSIM$\uparrow$ & LPIPS$\downarrow$ & FID$\downarrow$ & L1$\downarrow$ & RMSE$\downarrow$
		& SSIM$\uparrow$ & LPIPS$\downarrow$ & FID$\downarrow$ & L1$\downarrow$ & RMSE$\downarrow$ \\ 
            \hline 
		w/o $\mathcal{L}_{csh}$ &  0.619 &0.241 &110.32 &6.75 &3.98 &0.587 &0.301 &115.23 &7.51 &4.37\\
		w/o HAE &  0.621 &0.228 &109.37 &6.51 &3.86 &  0.606 &0.298 &111.25 &7.08 &4.15\\ 
		MX-Font++ & \textbf{0.689} & \textbf{0.201} & \textbf{103.94} & \textbf{6.03} & \textbf{3.23} & \textbf{0.653} & \textbf{0.279} & \textbf{108.37} & \textbf{6.58} & \textbf{3.49}\\ 
            \hline
	\end{tabular}
        }

\end{table}

Table~\ref{tab:ablation} shows that removing $\mathcal{L}_{csh}$ or HAE significantly impacts the model's performance. MX-Font++ (with both $\mathcal{L}_{csh}$ and HAE) outperforms the other configurations in metrics like SSIM, LPIPS, FID, L1, and RMSE. This highlights the importance of $\mathcal{L}_{csh}$ and HAE in enhancing the quality and accuracy of generated font images.

Thus, $\mathcal{L}_{csh}$ and HAE are crucial for the model to generate accurate and stylistically rich fonts. Retaining these components is essential for improving the quality and generalization of generated fonts.

\section{Conclusion}

This paper presents MX-Font++, a novel few-shot font generation method incorporating Heterogeneous Aggregation Experts and a content-style homogeneity loss. Our approach excels in cross-lingual font generation, particularly for low-resource languages, outperforming existing techniques in visual quality and metrics. MX-Font++ shows notable improvements for complex scripts like Chinese characters and demonstrates superior performance in user studies. Significantly, the synthetic fonts generated by MX-Font++ enhance the Scene Text Recognition model performance, advancing text recognition in low-resource language scenarios. This research marks a substantial improvement in font generation and its practical applications in multilingual image  processing systems.

\clearpage
\bibliographystyle{IEEEtran}
\bibliography{main}

\begin{thebibliography}{10}
\providecommand{\url}[1]{#1}
\csname url@samestyle\endcsname
\providecommand{\newblock}{\relax}
\providecommand{\bibinfo}[2]{#2}
\providecommand{\BIBentrySTDinterwordspacing}{\spaceskip=0pt\relax}
\providecommand{\BIBentryALTinterwordstretchfactor}{4}
\providecommand{\BIBentryALTinterwordspacing}{\spaceskip=\fontdimen2\font plus
\BIBentryALTinterwordstretchfactor\fontdimen3\font minus \fontdimen4\font\relax}
\providecommand{\BIBforeignlanguage}[2]{{%
\expandafter\ifx\csname l@#1\endcsname\relax
\typeout{** WARNING: IEEEtran.bst: No hyphenation pattern has been}%
\typeout{** loaded for the language `#1'. Using the pattern for}%
\typeout{** the default language instead.}%
\else
\language=\csname l@#1\endcsname
\fi
#2}}
\providecommand{\BIBdecl}{\relax}
\BIBdecl

\bibitem{tian2016rewrite}
Y.~Tian, ``Rewrite: Neural style transfer for chinese fonts, 2016,'' \emph{Retrieved Nov}, vol.~23, p. 2016, 2016.

\bibitem{tian2017zi2zi}
------, ``zi2zi: Master chinese calligraphy with conditional adversarial networks, 2017,'' \emph{Retrieved Jun}, vol.~3, p. 2017, 2017.

\bibitem{lyu2017auto}
P.~Lyu, X.~Bai, C.~Yao, Z.~Zhu, T.~Huang, and W.~Liu, ``Auto-encoder guided gan for chinese calligraphy synthesis,'' in \emph{2017 14th IAPR International Conference on Document Analysis and Recognition (ICDAR)}, vol.~1.\hskip 1em plus 0.5em minus 0.4em\relax IEEE, 2017, pp. 1095--1100.

\bibitem{azadi2018multi}
S.~Azadi, M.~Fisher, V.~G. Kim, Z.~Wang, E.~Shechtman, and T.~Darrell, ``Multi-content gan for few-shot font style transfer,'' in \emph{Proceedings of the IEEE conference on computer vision and pattern recognition}, 2018, pp. 7564--7573.

\bibitem{xie2024weakly}
Y.~Xie, X.~Chen, H.~Zhan, P.~Shivakumara, B.~Yin, C.~Liu, and Y.~Lu, ``Weakly supervised scene text generation for low-resource languages,'' \emph{Expert Systems with Applications}, vol. 237, p. 121622, 2024.

\bibitem{tang2022few}
L.~Tang, Y.~Cai, J.~Liu, Z.~Hong, M.~Gong, M.~Fan, J.~Han, J.~Liu, E.~Ding, and J.~Wang, ``Few-shot font generation by learning fine-grained local styles,'' in \emph{Proceedings of the IEEE/CVF conference on computer vision and pattern recognition}, 2022, pp. 7895--7904.

\bibitem{wang2023cf}
C.~Wang, M.~Zhou, T.~Ge, Y.~Jiang, H.~Bao, and W.~Xu, ``Cf-font: Content fusion for few-shot font generation,'' in \emph{Proceedings of the IEEE/CVF Conference on Computer Vision and Pattern Recognition}, 2023, pp. 1858--1867.

\bibitem{zheng2020jnd}
W.~Zheng, L.~Yan, C.~Gou, and F.-Y. Wang, ``Jnd-gan: human-vision-systems inspired generative adversarial networks for image-to-image translation,'' in \emph{ECAI 2020}.\hskip 1em plus 0.5em minus 0.4em\relax IOS Press, 2020, pp. 2816--2823.

\bibitem{chang2018pairedcyclegan}
H.~Chang, J.~Lu, F.~Yu, and A.~Finkelstein, ``Pairedcyclegan: Asymmetric style transfer for applying and removing makeup,'' in \emph{Proceedings of the IEEE conference on computer vision and pattern recognition}, 2018, pp. 40--48.

\bibitem{li2018beautygan}
T.~Li, R.~Qian, C.~Dong, S.~Liu, Q.~Yan, W.~Zhu, and L.~Lin, ``Beautygan: Instance-level facial makeup transfer with deep generative adversarial network,'' in \emph{Proceedings of the 26th ACM international conference on Multimedia}, 2018, pp. 645--653.

\bibitem{choi2018stargan}
Y.~Choi, M.~Choi, M.~Kim, J.-W. Ha, S.~Kim, and J.~Choo, ``Stargan: Unified generative adversarial networks for multi-domain image-to-image translation,'' in \emph{Proceedings of the IEEE conference on computer vision and pattern recognition}, 2018, pp. 8789--8797.

\bibitem{liu2019few}
M.-Y. Liu, X.~Huang, A.~Mallya, T.~Karras, T.~Aila, J.~Lehtinen, and J.~Kautz, ``Few-shot unsupervised image-to-image translation,'' in \emph{Proceedings of the IEEE/CVF international conference on computer vision}, 2019, pp. 10\,551--10\,560.

\bibitem{tumanyan2022splicing}
N.~Tumanyan, O.~Bar-Tal, S.~Bagon, and T.~Dekel, ``Splicing vit features for semantic appearance transfer,'' in \emph{Proceedings of the IEEE/CVF Conference on Computer Vision and Pattern Recognition}, 2022, pp. 10\,748--10\,757.

\bibitem{sun2017learning}
D.~Sun, T.~Ren, C.~Li, H.~Su, and J.~Zhu, ``Learning to write stylized chinese characters by reading a handful of examples,'' \emph{arXiv preprint arXiv:1712.06424}, 2017.

\bibitem{zhang2018separating}
Y.~Zhang, Y.~Zhang, and W.~Cai, ``Separating style and content for generalized style transfer,'' in \emph{Proceedings of the IEEE conference on computer vision and pattern recognition}, 2018, pp. 8447--8455.

\bibitem{gao2019artistic}
Y.~Gao, Y.~Guo, Z.~Lian, Y.~Tang, and J.~Xiao, ``Artistic glyph image synthesis via one-stage few-shot learning,'' \emph{ACM Transactions on Graphics (TOG)}, vol.~38, no.~6, pp. 1--12, 2019.

\bibitem{liu2023fonttransformer}
Y.~Liu and Z.~Lian, ``Fonttransformer: Few-shot high-resolution chinese glyph image synthesis via stacked transformers,'' \emph{Pattern Recognition}, vol. 141, p. 109593, 2023.

\bibitem{bai2024intelligent}
Y.~Bai, Z.~Huang, W.~Gao, S.~Yang, and J.~Liu, ``Intelligent artistic typography: A comprehensive review of artistic text design and generation,'' \emph{arXiv preprint arXiv:2407.14774}, 2024.

\bibitem{park2021multiple}
S.~Park, S.~Chun, J.~Cha, B.~Lee, and H.~Shim, ``Multiple heads are better than one: Few-shot font generation with multiple localized experts,'' in \emph{Proceedings of the IEEE/CVF international conference on computer vision}, 2021, pp. 13\,900--13\,909.

\bibitem{cha2020few}
J.~Cha, S.~Chun, G.~Lee, B.~Lee, S.~Kim, and H.~Lee, ``Few-shot compositional font generation with dual memory,'' in \emph{Computer Vision--ECCV 2020: 16th European Conference, Glasgow, UK, August 23--28, 2020, Proceedings, Part XIX 16}.\hskip 1em plus 0.5em minus 0.4em\relax Springer, 2020, pp. 735--751.

\bibitem{jacobs1991adaptive}
R.~A. Jacobs, M.~I. Jordan, S.~J. Nowlan, and G.~E. Hinton, ``Adaptive mixtures of local experts,'' \emph{Neural computation}, vol.~3, no.~1, pp. 79--87, 1991.

\bibitem{du2022svtr}
Y.~Du, Z.~Chen, C.~Jia, X.~Yin, T.~Zheng, C.~Li, Y.~Du, and Y.-G. Jiang, ``Svtr: Scene text recognition with a single visual model,'' \emph{arXiv preprint arXiv:2205.00159}, 2022.

\bibitem{liao2019scene}
M.~Liao, J.~Zhang, Z.~Wan, F.~Xie, J.~Liang, P.~Lyu, C.~Yao, and X.~Bai, ``Scene text recognition from two-dimensional perspective,'' in \emph{Proceedings of the AAAI conference on artificial intelligence}, vol.~33, no.~01, 2019, pp. 8714--8721.

\bibitem{wan2020textscanner}
Z.~Wan, M.~He, H.~Chen, X.~Bai, and C.~Yao, ``Textscanner: Reading characters in order for robust scene text recognition,'' in \emph{Proceedings of the AAAI conference on artificial intelligence}, vol.~34, no.~07, 2020, pp. 12\,120--12\,127.

\bibitem{xie2024deeptts}
Y.~Xie, C.~Xu, C.~Shi, J.~Li, Z.~Yuan, and Q.~Qiao, ``Deeptts: Enhanced transformer-based text spotter via deep interaction between detection and recognition tasks,'' in \emph{Pacific Rim International Conference on Artificial Intelligence}.\hskip 1em plus 0.5em minus 0.4em\relax Springer, 2024, pp. 450--462.

\bibitem{xie2024dntextspotter}
Y.~Xie, Q.~Qiao, J.~Gao, T.~Wu, J.~Fan, Y.~Zhang, J.~Zhang, and H.~Sun, ``Dntextspotter: Arbitrary-shaped scene text spotting via improved denoising training,'' \emph{arXiv preprint arXiv:2408.00355}, 2024.

\bibitem{wan2020vocabulary}
Z.~Wan, J.~Zhang, L.~Zhang, J.~Luo, and C.~Yao, ``On vocabulary reliance in scene text recognition,'' in \emph{Proceedings of the IEEE/CVF Conference on Computer Vision and Pattern Recognition}, 2020, pp. 11\,425--11\,434.

\bibitem{sun2024haformer}
D.~Sun, Y.~Wang, J.~Z. Zuo, and H.~Zheng, ``Haformer: Heterogeneous aggregation transformer for single image deraining,'' in \emph{ICASSP 2024-2024 IEEE International Conference on Acoustics, Speech and Signal Processing (ICASSP)}.\hskip 1em plus 0.5em minus 0.4em\relax IEEE, 2024, pp. 6050--6054.

\bibitem{yim2021synthtiger}
M.~Yim, Y.~Kim, H.-C. Cho, and S.~Park, ``Synthtiger: Synthetic text image generator towards better text recognition models,'' in \emph{International conference on document analysis and recognition}.\hskip 1em plus 0.5em minus 0.4em\relax Springer, 2021, pp. 109--124.

\bibitem{zhang2018unreasonable}
R.~Zhang, P.~Isola, A.~A. Efros, E.~Shechtman, and O.~Wang, ``The unreasonable effectiveness of deep features as a perceptual metric,'' in \emph{Proceedings of the IEEE conference on computer vision and pattern recognition}, 2018, pp. 586--595.

\bibitem{heusel2017gans}
M.~Heusel, H.~Ramsauer, T.~Unterthiner, B.~Nessler, and S.~Hochreiter, ``Gans trained by a two time-scale update rule converge to a local nash equilibrium,'' \emph{Advances in neural information processing systems}, vol.~30, 2017.

\bibitem{kong2022look}
Y.~Kong, C.~Luo, W.~Ma, Q.~Zhu, S.~Zhu, N.~Yuan, and L.~Jin, ``Look closer to supervise better: One-shot font generation via component-based discriminator,'' in \emph{Proceedings of the IEEE/CVF conference on computer vision and pattern recognition}, 2022, pp. 13\,482--13\,491.

\bibitem{yang2024fontdiffuser}
Z.~Yang, D.~Peng, Y.~Kong, Y.~Zhang, C.~Yao, and L.~Jin, ``Fontdiffuser: One-shot font generation via denoising diffusion with multi-scale content aggregation and style contrastive learning,'' in \emph{Proceedings of the AAAI Conference on Artificial Intelligence}, vol.~38, no.~7, 2024, pp. 6603--6611.

\end{thebibliography}

\end{document}